\pdfoutput=1

\documentclass[11pt]{article}

\usepackage{emnlp2021}

\usepackage{times}
\usepackage{bbm}
\usepackage{graphicx}
\usepackage{latexsym}
\usepackage{arydshln}
\usepackage{dblfloatfix}

\usepackage[T1]{fontenc}

\usepackage[utf8]{inputenc}

\usepackage{microtype}
\usepackage{amssymb,amsmath}
\title{Not All Negatives are Equal: \\ Label-Aware Contrastive Loss for Fine-grained Text Classification}

\author{Varsha Suresh \\
  Dept. of Computer Science \\
  National University of Singapore \\
  \texttt{varshasuresh@u.nus.edu} \\\And
  Desmond C. Ong \\
  Dept. of Information Systems and Analytics \\
  National University of Singapore, \\ 
  \& Institute of High Performance \\ Computing, A*STAR \\
  \texttt{dco@comp.nus.edu.sg} \\}

\begin{document}
\maketitle
\begin{abstract}

Fine-grained classification involves dealing with datasets with larger number of classes with subtle differences between them. Guiding the model to focus on differentiating dimensions between these commonly confusable classes is key to improving performance on fine-grained tasks. In this work, we analyse the contrastive fine-tuning of pre-trained language models on two fine-grained text classification tasks, emotion classification and sentiment analysis. We adaptively embed class relationships into a contrastive objective function to help differently weigh the positives and negatives, and in particular, weighting closely confusable negatives more than less similar negative examples. We find that Label-aware Contrastive Loss outperforms previous contrastive methods, in the presence of larger number and/or more confusable classes, and helps models to produce output distributions that are more differentiated. 

\end{abstract}

\section{Introduction}

Fine-grained classification involves distinguishing between classes that have subtle variations among them. For example, in image classification, we can classify birds from non-birds, or attempt a more fine-grained classification of bird species \cite{akata2015evaluation}. In NLP, one example is sentiment analysis, where we could have a coarse positive/negative classification, or a fine-grained set of categories that differentiate ``positive'' and ``very positive'' (i.e., an ordinal scale), such as in \citet{socher2013recursive}. Similarly, for emotion classification, we could try to classify a text into 4 to 6 emotions, or into much finer classifications of 27 \cite{demszky2020goemotions} or 32 \cite{rashkin2019towards} emotion categories. This involves distinguishing between some closely confusable pairs of emotions, such as ``sad'' and ``devastated'', or ``furious'' and ``annoyed''. Fine-grained classification tasks are challenging precisely due to the presence of class interference amongst closely confusable classes \cite{collins2018evolutionary,zhao2017survey}.


The standard approach today to task classification involves using a pre-trained language model (e.g., BERT) which is fine-tuned on downstream tasks using a standard cross-entropy loss. However, this standard loss may not be the optimal manner in which to train fine-grained classification models. A simple counterexample is that cross-entropy loss treats misclassifications as nominal, not ordinal, so misclassifying a ``positive'' as a ``very positive'' is no worse (in terms of the loss) as ``very negative''. But even within nominal categories, misclassifying ``annoyed'' as ``furious'' is quite different from a misclassification of ``joyful'', as there are varying degrees of semantic similarity between nominal categories. Intuitively, we can try to improve model performance by modifying the loss to reflect the \emph{contrast} between pairs of examples of the same or different classes. Such contrastive approaches are widely used in computer vision tasks for label-noise reduction, semi-supervised and self-supervised learning tasks \cite{le2020contrastive}. More recently in NLP, \citet{gunel2021supervised} used a supervised contrastive loss to improve fine-tuning performance of pre-trained language models in several few-shot learning scenarios.



In this work, we incorporate inter-class relationships into a Label-aware Contrastive Loss (LCL), which helps the model to differentiate the weights between different negative samples. At a high level, the model adaptively learns which pairs of classes are more similar, and which are more different. We use a dual-model approach where a weighting model learns the inter-label relationships that are used in the main embedding model's contrastive objective. We evaluate our approach on two popular tasks in NLP: emotion recognition (4 datasets to span both coarse- and fine-grained classification), and sentiment analysis (with a coarse and fine-grained version of the same dataset). We find that LCL outperforms existing contrastive learning losses, and performs comparably with the state-of-the-art. We supplement our findings with targeted experiments to provide evidence for boundary conditions---situations in which LCL should work best---and for how LCL affects model prediction confidence.


\section{Related Work}

\subsection{Fine-grained classification}

Fine-grained classification is a popular problem in image classification, including tasks like distinguishing between different animal species \cite{wei2019deep, zhao2017survey}. We note that in NLP, ``fine-grained'' is commonly used when analysing different granularities of text, such as character-, word- and span-level information \cite{zirn2011fine,da2019fine,liu2020fine}. In this work, we use fine-grained classification to refer to the \emph{nature of labels associated with the task}.

Fine-grained classification tasks involve finding subtle differences to distinguish between close classes. For instance, ``coarse" sentiment classification involves distinguishing negative and positive sentiments in text, and fine-grained sentiment classification involves further distinguishing the \emph{positive} class into \emph{very positive} and \emph{positive}. This problem is challenging because the classes are semantically similar, which makes it difficult for the model to learn the labels \cite{collins2018evolutionary}.

Recent models have applied state-of-the-art attention mechanisms and multi-task learning to solve fine-grained sentiment classification. \citet{balikas2017multitask} performed fine-grained sentiment classification using a multi-task learning setup that performed both binary and fine-grained sentiment classification simultaneously. \citet{yin2020sentibert} composed the sentiment semantics using an attention network to enhance BERT's pre-training objective, and showed improvement in a downstream fine-grained sentiment analysis task. \citet{tian2020skep} modified the pre-training objectives of language models to include more sentiment-specific tasks, such as sentiment word masking and sentiment word prediction, and showed improved performance in fine-grained sentiment analysis. 
These previous methods mostly focus on improving the pre-training of language models, or incorporating multiple task training; here, we focus on improving contrastive fine-tuning to solve fine-grained text classification.


Another important fine-grained classification task is that of emotion recognition. Traditionally, emotion recognition datasets have a small number of emotions (e.g., 4-7). Two recent datasets were proposed to address this issue: 
\citet{rashkin2019towards} introduced Empathetic Dialogues, which contains text conversations labelled with 32 emotion labels, and \citet{demszky2020goemotions} introduced GoEmotions, which contains Reddit comments labelled with 27 emotion labels. Recently, \citet{suresh2021knowledge} introduced a method to incorporate knowledge from emotion lexicons into an attention mechanism to improve fine-grained emotion classification on these two datasets. \citet{khanpour2018fine} similarly used lexicon-based features to tackle fine-grained emotion recognition from online health posts. However, there is still much work to be done in fine-grained emotion classification, and it has important implications for designing empathetic agents and chatbots \cite{roller2021recipes}.


Finally, we note that fine-grained classification has also been explored in the context of entity-type classification \cite{ling2012fine,jin2019fine}. However, this task is generally multi-label in nature and is out of the scope of the current work.

\subsection{Contrastive learning}


Contrastive learning focuses on improving the ability of the model to differentiate a given data point from ``positive'' examples (points sharing the same label) and from ``negative'' examples (different labels).
Contrastive learning has been widely used in computer vision, especially in self-supervised settings \cite{le2020contrastive,chen2020simple} where such learning guides the model based on similarities between the latent representation of the samples. \cite{chen2020simple} introduced SimCLR, a simplified version of contrastive loss that does not use memory banks \cite{tian2020contrastive,he2020momentum,misra2020self} or designated architectures \cite{bachman2019learning}, and which achieves improved performance in both semi-supervised and self-supervised settings. SimCLR uses data augmentation to create ``positive'' examples that are similar to a given input.
\citet{khosla2020supervised} extended SimCLR to also leverage label information: they include other training examples with the same label in the set of ``positive'' examples.


Contrastive loss has also been recently incorporated in both the pre-training and fine-tuning objectives of pre-trained language models. Self-supervised contrastive loss has been used for pre-training language models such as BERT \cite{fang2020cert,meng2021coco}. \citet{gunel2021supervised} used a combination of cross entropy and supervised contrastive loss for fine-tuning pre-trained language models to improve performance in few-shot learning scenarios. \citet{gao2021simcse} used a contrastive objective to fine-tune pre-trained language models to obtain sentence embeddings, and achieved state-of-the-art performance in sentence similarity tasks. In our work, we aim to improve the fine-tuning objective of pre-trained language models for downstream tasks involving fine-grained classes. 
 
\subsection{Other related work}

In addition to the above works, we mention other related references which used similar techniques.
Dual-model approaches are used in tasks like knowledge distillation, where the knowledge from a larger \emph{teacher} network is transferred to a lighter \emph{student} model \cite{hinton2015distilling,kim2016sequence,sun2020contrastive,sun2019patient,li2020bert,aguilar2020knowledge}, however, these works are mainly focused on model compression. 
Dual-model strategies have also been widely used in label-noise representation learning in image classification tasks \cite{han2018co,wei2020combating,lu2021co,feng2019learning} by updating each other with \textit{clean} samples (the samples which have the lowest loss value in every iteration). 
However, the sample selection performed by these works assume that the noise rate in each dataset is known or needs to be estimated, which is not always possible.

Another set of works focus on sample re-weighting to focus on select samples more. \citet{plank2014learning} use inter-annotator agreement to guide the model's focus on samples that are harder to distinguish. Sample re-weighting is also widely used to reduce label noise. Although the majority of works in this area depend on a pre-determined weighting function, there are a few notable papers which automate this process by adaptively calculating weights: \citet{chang2017active} uses active learning to re-weight samples, while \citet{ren2018learning} uses gradients to learn weights, however their performance drops with large number of classes \cite{song2020learning}. Meta-Weight-Net uses a single-layer neural network to obtain the weights \cite{shu2019meta}. These methods all require clean validation data to optimize their learning objective.

\section{Approach}
\begin{figure*}[h]
\includegraphics[width=0.9\textwidth]{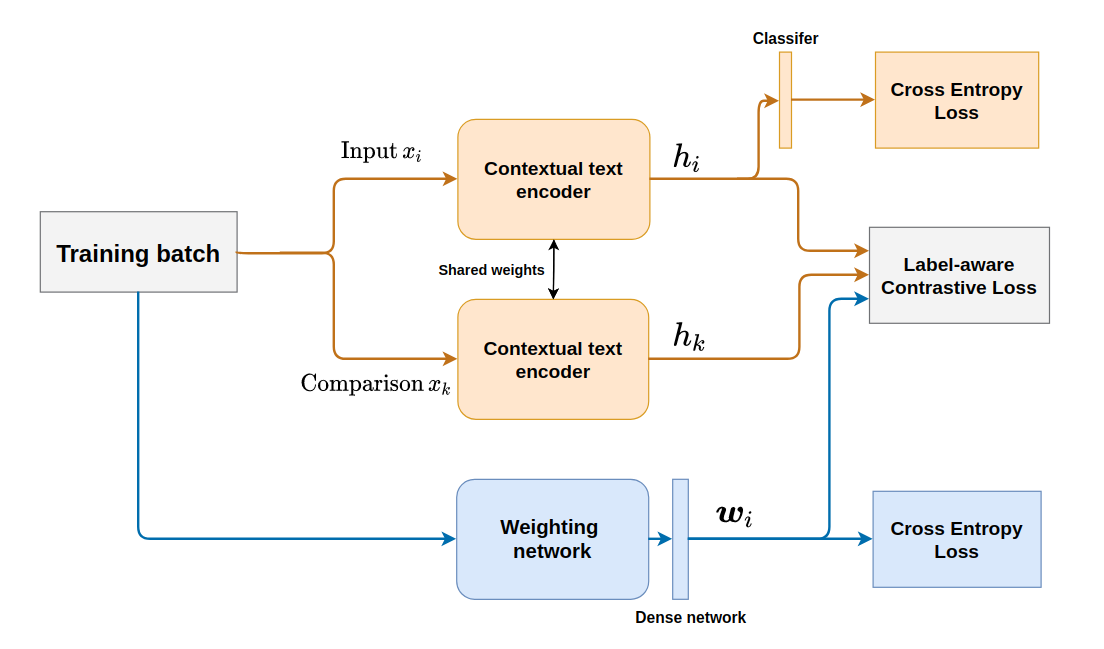}
\centering
\caption{Illustration of training strategy used in our Label-aware Contrastive Loss approach. The encoder network is in orange and the the weighting network is indicated in blue. In the encoder network, every sample from the training batch is compared against every other sample in the Label-aware Contrastive Loss function. Note that at testing time, only the contextual encoder is used. }
\label{fig:model}
\end{figure*}

\subsection{Contrastive Loss}
A Contrastive Loss (CL) brings the latent representations of samples belonging to the same class closer together, by defining a set of positives (that should be closer) and negatives (that should be further apart). 
The type of positives and negatives vary and is dependent on the contrastive loss used. Throughout this section we denote the set of positives as $\mathcal{P}$ and set of negatives as $\mathcal{N}$. Let us also denote a batch of sample and label pairs as $\{x_{i},y_{i}\}_{i \in I}$,
where $I = \{1, \cdots,K\}$ is the indices of the samples and $K$ is the batch-size. 

In the self-supervised version of contrastive loss \cite{chen2020simple}, one applies augmentation to all $K$ samples to produce $K$ augmented data-points. 
Therefore, the batch size becomes $2K$ and $I = \{1, \cdots, 2K\}$. The positive set for a given $x_{i}$ contains only one sample, the augmented version of $x_{i}$, and we denote its index as $g(i)$. The negative set would be the rest of the samples in the batch. The loss is defined as:

\begin{equation}
L_{self} = \sum_{i=1}^{2K} -\log \frac{\exp(h_{i} \cdot h_{g(i)}/\tau)}{\sum_{k \in \mathcal{I}/i} \exp(h_{i} \cdot h_{k}/\tau)}
\end{equation}
where $\tau$ is the temperature hyper-parameter. Larger values of $\tau$ scale down the dot-products, creating more difficult comparisons. 
$h_{i}$ is the normalised representation vector of $x_{i}$ obtained from an encoder $\Phi$.

\citet{khosla2020supervised} extended the above loss to a Supervised Contrastive Loss (SCL) by including the samples belonging to the same class as $x_{i}$ in its positive set. The positive set is given by $\mathcal{P} = \{p : {p \in I} , {y_{p} = y_{i}} \land p \neq i\}$, with size $|\mathcal{P}|$. The supervised contrastive loss is given by: 
\begin{equation}
\label{eq:scl}
L_{SCL} = \sum_{i=1}^{2K} \frac{-1}{|\mathcal{P}|} \sum_{p \in \mathcal{P}} \log \frac{\exp(h_{i} \cdot h_{p}/\tau)}{\sum_{k \in \mathcal{I}/i} \exp(h_{i} \cdot h_{k}/\tau)}
\end{equation}

\subsection{Label-aware Contrastive Loss}

In our work, we introduce relationships between class labels to adaptively distinguish between the negative examples.  
From Eqn. \ref{eq:scl} we can see that Supervised Contrastive Loss weights all positive and negative samples equally to the current sample $x_{i}$. But not all negatives are equal. In certain fine-grained text classification tasks, we have semantically-similar labels with more subtle differences, and are thus more \textit{confusable}. For example, ``sad'' and ``devastated'' are semantically closer emotion categories than ``sad'' and ``happy''. Thus, our goal was to introduce a method for adaptively weighting a given input's positive/negative samples based on the label-relationships between them, thereby helping the model differentiate the more difficult negatives.

We propose Label-aware Contrastive Loss (LCL) which adapts Contrastive Loss for fine-grained classification tasks by incorporating inter label-relationships. For the positive set, we follow \cite{khosla2020supervised, gunel2021supervised} where $\mathcal{P}$ of a given sample contains the augmented sample and samples within the same class. We utilise a weighting vector $\boldsymbol{w_{i}} \in \mathbb{R}^C$ where $C$ is total number of classes to weight the pair-wise similarity values of the supervised contrastive loss defined in Eq. \ref{eq:scl}. Our adapted loss function for each entry $i$ and total across the batch is: 
\begin{equation}
    \mathcal{L}_{i} = \sum_{p \in \mathcal{P}} \log \frac{ w_{i,y_{i}} \cdot \exp(h_{i} \cdot h_{p}/\tau)}{\sum_{k \in \mathcal{I} \setminus i} w_{i,y_{k}} \cdot \exp(h_{i} \cdot h_{k}/\tau)} 
\end{equation}

\begin{equation}
L_{LCL} = \sum_{i=1}^{2K} \frac{-1}{|\mathcal{P}|}\mathcal{L}_{i} \label{eq:lcl}
\end{equation}

Here, $w_{i,y_{k}}$ indicates the relationship between an input $x_{i}$ and a label $y_{k}$. Just as in the previous losses, $h_{i} \in \mathbb{R}^{d}$ is the output representation of the encoder for $x_{i}$. We normalise $h_{i}$ for the similarity comparison, similar to \citet{chen2020simple}. 

In contrastive loss we want the weights of the positives to be higher and that of the negatives to be lower. However, we want to increase the weight of confusable negative labels relative to other negative labels. 
In our work, we aim to incorporate these inter-label relationships into the contrastive objective. To weigh each comparison sample differently, in addition to a primary encoder $\Phi$, we use a weighting network $\Psi$. 
We follow a dual-model strategy similar to co-teaching approaches \cite{han2018co,wei2020combating} where the weighting network is a second network that coordinates with the primary encoder. 
The input batch is fed into $\Psi$ and output is optimised using Cross-entropy loss $L_{w}$. The prediction probabilities obtained from the softmax layer, i.e. soft labels, is used to obtain confidence of the current sample, is given by: 
\begin{equation}
\label{eq:weight}
\boldsymbol{w_{i}}= \frac{\exp(h_{i})}{\sum_{c=1}^{C} \exp(h_{i})} 
\end{equation}
Here, $\boldsymbol{w_{i}} = \{w_{i,c}\}_{c=1}^{C}$ where $C$ is the total number of classes. Each $w_{i,c}$ denotes the confidence of the weighting network that sample $x_{i}$ belongs to class $c$. 
When $\Psi$ is given a \textit{confusable} sample, it will have higher scores for the classes that are more closely associated with the current sample. We hypothesize that incorporating these high values back into the negative comparison in the supervised contrastive loss of the primary encoder would steer the encoder toward finding more distinguishing patterns to differentiate between \textit{confusable} samples.

\textbf{Training setup:} The output vector of the weighting network is optimized using a Cross Entropy Loss $L_{w}$, while the output of the encoder network is optimized by using a linear combination of $L_{LCL}$ and Cross Entropy Loss $L_{e}$. The encoder and weighting networks are jointly optimised using objective function $L_{f}$: 
\begin{equation}
    L_{f} = \alpha (L_{w} + L_{e})  + (1-\alpha) L_{LCL}
\end{equation}
Here, $\alpha$ is a tunable loss scaling factor similar to \citet{gunel2021supervised}. We note that both the encoder and the weighting network are utilised during training, but in the testing phase, we use only the primary encoder network.

The overall training process is shown in Fig. \ref{fig:model}. Each input training batch $I$ is passed to the encoder network $\Phi$ and the weighting network $\Psi$ simultaneously. Here, both these networks are initialised by a pre-trained language model and the [CLS] token of the last layer of $\Phi$ is the final representation $h_{i}$ which is used for computing $L_{LCL}$. For performing the classification, $h_{i}$ is projected down using the classifier and the output is optimised using cross-entropy loss $L_{e}$. The architecture of the weighting network was designed in the same way as the fine-tuning setup of the pre-trained language model of choice, and the weight vector $\boldsymbol{w}_{i}$ is the output probability vector obtained after the softmax projection.

\section{Experiments}

\subsection{Datasets}
We evaluate our approach using two tasks, Emotion Recognition and Sentiment Analysis. We choose these tasks as it helps demonstrate our model’s performance in different types of inter-class relationships that exist in text classification. Specifically, in sentiment classification the classes are ordinal, whereas in emotion recognition the classes are nominal\footnote{Although there still may be underlying latent structure such that some classes may be semantically more similar than others, e.g., \emph{afraid} vs. \emph{anxious} vs. \emph{joyful}.}.

For emotion recognition, we use the following 4 datasets, ordered in decreasing number of classes:



\begin{itemize}
    \item Empathetic Dialogues \cite{rashkin2019towards}\footnote[2]{\url{https://github.com/facebookresearch/EmpatheticDialogues}}: a dataset of two-way conversations between a speaker and listener, and labelled with 32 emotions. In this work, we only use the first turn of the conversation, which consists of the speaker describing an emotional incident. The train/validation/test split for the dataset is 19,533 / 2,770 / 2,547 samples respectively.
    \item GoEmotions \cite{demszky2020goemotions}\footnote[3]{\url{https://github.com/google-research/google-research/tree/master/goemotions}}, a dataset of Reddit comments labelled with 27 emotions (we did not include samples with neutral label). The original dataset is multi-labelled, i.e, some samples have more than one label. In this work, we use only the single-labelled samples, which is $\sim$80\% of the total data. The train/validation/test split of this dataset is 23,485 / 2,956 / 2,984.
    \item ISEAR (International Survey on Emotion Antecedents and Reactions) \cite{scherer1994evidence}\footnote[4]{\url{https://www.unige.ch/cisa/research/materials-and-online-research/research-material/}} contains sentences of emotion experiences labelled with one of 7 emotion categories. The train/validation/test split of the dataset is 4,599 / 1,533 / 1,534. 
    \item EmoInt \cite{mohammad2017wassa}\footnote[5]{\url{http://saifmohammad.com/WebPages/EmotionIntensity-SharedTask.html}} consists of tweets labelled with one of 4 emotion categories. The train/validation/test split of this dataset is 3,612 / 346 / 3,141.
\end{itemize}





For Sentiment Analysis, we use the 5-class and 2-class classification versions of the Standford Sentiment Treebank \cite{socher2013recursive}, which consists of movie reviews annotated for sentiment. The SST-5 has 5 classes (\emph{very negative}, \emph{negative}, \emph{neutral}, \emph{positive}, and \emph{very positive}), while the SST-2 is only a binary (\emph{negative}/\emph{positive}) classification. The train/validation/test split for the SST-5 is 8,544/ 1,101 / 2,210, and for SST-2 is 6,920 / 872 / 1,821.

\subsection{Implementation Details}

We initialised both the pre-trained encoder and weighting network using  $\text{ELECTRA}_{base}$ (\texttt{electra-base-discriminator}) from HuggingFace’s Transformers library \cite{wolf2019transformers}, which consists of 12 Transformer layers with a hidden representation size of 768. As is convention, we use the representation corresponding to the [CLS] token of the last layer as an input into the final classification layer \cite{clark2019electra}. The classifier present in the primary encoder consists of a 2-layer dense network with the first layer having hidden size of 768 with a ReLU activation, followed by an output layer. The dropout was set to 0.1. 


Similar to previous research \cite{khosla2020supervised,gunel2021supervised}, we use data augmentation to generate positive samples. Here, we use synonym replacement where we substitute 30\% of the words in the input text by replacing it with words with semantic similarity using WordNet dictionary \cite{miller1995wordnet}. The coverage of the WordNet dictionary was $\sim$69\% for EmpatheticDialogues, $\sim$69\% for SST-2 and SST-5, $\sim$66\% for ISEAR, $\sim$62\% for EmoInt and $\sim$61\% for GoEmotions. Previous research \cite{wei2019eda} have shown that synonym replacement works well as it could introduce new vocabulary words and help the model generalise. In addition, synonym replacement does not require an external model unlike other augmentation methods like back-translation. 

For training, we used the Adam optimiser and early stopping based on performance on the validation set. We ran our models with 5 random seed settings and report the mean performance. More details regarding the hyper-parameter settings and computing infrastructure can be found in the Appendix. Source code is available at \url{https://github.com/varsha33/LCL_loss}.

\subsection{Model comparisons and evaluation}


For the emotion classification task we calculate classification accuracy and F1 score, while for sentiment analysis we compare accuracy of sentence-level sentiment classification.
For both tasks, we compare LCL against the following baselines:
\begin{itemize}
    \item \textbf{Fine-tuning objectives}: We compare against the standard Cross-entropy Loss, as well as Supervised Contrastive Loss (SCL) \cite{gunel2021supervised}. In both comparisons and in LCL, we use $\text{ELECTRA}_{base}$ as the pre-trained language model.
    \item \textbf{General pre-trained language models}: For emotion classification, we also compare with $\text{BERT}_{base}$ \cite{devlin2019bert} as our baseline. We use the same fine-tuning architecture as \citet{devlin2019bert}. For sentiment analysis, we compare against $\text{BERT}_{base}$  (SST-2 \cite{devlin2019bert} and SST-5 \cite{munikar2019fine}) and $\text{RoBERTa}_{base}$ \cite{liu2019roberta}.
    \item \textbf{Sentiment-specific language models}: For sentiment analysis, we compare against SentiBERT \cite{yin2020sentibert}, SentiLARE \cite{ke2019sentilare} and SKEP \cite{tian2020skep}, which are language models designed specifically for sentiment analysis and related tasks. 
\end{itemize}

\section{Results and Discussion}

\begin{table*}[htb]
\resizebox{\textwidth}{!}{%
\begin{tabular}{lcccc|cccc}
\hline
Dataset: & \multicolumn{2}{c}{Empathetic Dialogues} & \multicolumn{2}{c|}{GoEmotions} & \multicolumn{2}{c}{ISEAR} & \multicolumn{2}{c}{EmoInt} \\
Number of classes: & \multicolumn{2}{c}{32} & \multicolumn{2}{c|}{27} & \multicolumn{2}{c}{7} & \multicolumn{2}{c}{4} \\
& Acc / \% & F1 & Acc / \% & F1 & Acc / \% & F1 & Acc / \% & F1 \\ \hline
$\text{BERT}_{base}$ &
  55.8 (0.8) &
  54.4 (1.2) &
  64.1 (0.5) &
  63.0 (0.9) &
  69.2 (0.3) &
  69.3 (0.1) &
  85.0 (0.6) &
  85.0 (0.6) \\
\hline
ELECTRA$_{base}$ + Cross-Entropy Loss & 58.3 (0.5)         & 56.8 (0.5)        & 64.8 (0.3)                  & 63.9 (0.4)          & 71.4 (0.2) & 71.4 (0.2) & 85.5 (0.9)  & 85.5 (0.9)  \\
ELECTRA$_{base}$ + SCL \cite{gunel2021supervised} & 58.5 (0.7)  & 57.0 (0.9)        & 64.3 (0.4)                  & 63.0 (0.4)           & 70.5 (0.5) & 70.5 (0.6) & 85.7 (0.2)  & 85.8 (0.2) \\
ELECTRA$_{base}$ + LCL & \textbf{60.1 (0.3)} & \textbf{59.1 (0.3)} & \textbf{65.5 (0.2)} & \textbf{64.8 (0.2)} 
       & \textbf{72.4 (0.2)}         & \textbf{72.4 (0.2)}   & \textbf{86.6 (0.3)}         & \textbf{86.6 (0.3)} \\ \hline
\end{tabular}
}
\caption{Summary of results for fine-grained emotion recognition. We divide the table into fine-grained (left) and coarse-grained (right) emotion classification, based on the number of classes. We compare the results of an ELECTRA encoder trained with: a standard cross-entropy loss, a Supervised Contrastive Loss (SCL), and our proposed Label-aware Contrastive Loss (LCL). The results shown are averaged over 5 runs, with standard deviations in parenthesis. 
}
\label{tab:emo}
\end{table*}

\subsection{Emotion Classification Performance}
\label{res:emotion}

For emotion classification we compared our proposed Label-aware Contrastive Loss (LCL) work with the standard training objective, i.e., cross-entropy loss. We also compared with \citet{gunel2021supervised}'s formulation of Supervised Contrastive Loss (SCL), who used a linear combination of SCL and Cross-entropy loss for fine-tuning pre-trained language models (in contrast to the original SCL paper, \citealp{khosla2020supervised}, who used a two-stage training regime). For all fine-tuning objectives, we used $\text{ELECTRA}_{base}$ as the pre-trained language model. To evaluate the approaches we use top-1 Accuracy and weighted macro F1-score.

As shown in Table \ref{tab:emo}, our LCL objective function improved classification performance compared to both SCL and cross-entropy loss, on both fine-grained emotion classification (32-class, LCL$>$SCL, t-test on accuracy, $t=4.20, p=.007$, LCL$>$CEL, $t=6.42, p<.001$; and 27-class classification; LCL$>$SCL, $t=5.70, p<.001$, LCL$>$CEL, $t=4.32, p=.002$), as well as coarse-grained emotion classification (7-class, LCL$>$SCL, $t=7.39, p<.001$, LCL$>$CEL, $t=7.70, p<.001$; and 4-class classification, LCL$>$SCL, $t=5.34, p<.001$, LCL$>$CEL, $t=2.25, p=.078$ not significant). The consistent improved performance of LCL is in contrast to SCL, which did not outperform standard cross-entropy loss, (all $p>.05$, with SCL in fact performing worse than CEL on ISEAR, $t=3.34, p=.02$). 
These results suggest that incorporating class relationships into the fine-tuning objective of pre-trained language models can improve classification accuracies.

\subsection{Sentiment Analysis Performance}

\begin{table}[htb]
\centering
\resizebox{\columnwidth}{!}{%
\begin{tabular}{lcc}
\hline\hline
             & SST-5 & SST-2 \\
             & Acc / \% & Acc / \% \\ \hline
$\text{BERT}_{base}$ \cite{munikar2019fine} & 53.2 (-) & \\
$\text{BERT}_{base}$\cite{devlin2019bert}  & & 93.5(-) \\
$\text{RoBERTa}_{base}$ \cite{liu2019roberta} & 56.2 (-) & 94.8(-) \\
SentiBERT\cite{yin2020sentibert} &  57.8 (-) & 94.7 (-) \\
SentiLARE\cite{ke2019sentilare} &  \textbf{58.6 (-)} & \\
SKEP \cite{tian2020skep} & & \textbf{96.7 (-)}\\
$\text{ELECTRA}_{base}$\cite{clark2019electra}  & & 93.4 (-) \\
\hdashline
$\text{ELECTRA}_{base}$ (Our implementation) & 57.1 (1.2) & 94.4 (0.3)         \\
$\text{ELECTRA}_{base}$+ SCL \cite{gunel2021supervised} & 57.4 (0.6)  & 94.3 (0.2)       \\
$\text{ELECTRA}_{base}$ + LCL (Ours) & \textbf{58.5 (0.2)} & \textbf{94.5 (0.1)}   \\ \hline
\end{tabular}%
}
\caption{Summary of results for fine-grained (5-class) and coarse-grained (2-class) sentiment analysis. The results shown are averaged over 5 runs, with standard deviations in parenthesis.}
\label{tab:sentiment-results}
\end{table}

For sentiment analysis, we used the sentence inputs from SST-5 and SST-2. In addition to comparing LCL with varying fine-tuning objectives (cross-entropy and SCL), we also compare against recent state-of-the-art works, focusing on pre-trained language models and pre-trained language models learnt specifically for sentiment classification such as SentiBERT \cite{yin2020sentibert}, SentiLARE \cite{ke2019sentilare}, and SKEP \cite{tian2020skep}. To ensure a fair comparison, we use the $base$ version of the pre-trained language models unless mentioned otherwise. To evaluate, we use top-1 Accuracy.

From the results in Table \ref{tab:sentiment-results}, in the case of SST-5, our LCL objective showed improved classification performance compared to SCL ($t=3.61, p=.01$), and standard cross-entropy loss (SST-5: $t=2.40, p=.069$, although this is not significant due to high SD in CEL performance). 
Our LCL-fine-tuned model also achieves a performance comparable to the state-of-the-art performance of SentiLARE, although not statistically different ($p=.77$).
On SST-2, our LCL performance gains compared to cross-entropy and SCL are far more modest (neither were statistically significant; $p=.78$ and $p=.32$ respectively), and it performs comparably to previous SOTA pre-trained models, although it does not do as well as SKEP ($p<.001$). We provide two possible reasons: one, there is already very high performance (e.g. 94\% accuracies) on this binary classification task, which makes it difficult to get clear consistent improvements. Second and more importantly, we designed LCL to increase inter-class contrast, and so our method should work better for higher number of classification, compared to binary classification. Indeed, we see that LCL's improvements are much stronger and consistent on the fine-grained (5-class) sentiment classification task.


\subsection{Case Study: Varying number of classes}


\begin{table*}[]
\centering
\resizebox{\textwidth}{!}{%
\begin{tabular}{lccc|c|cccc}
Number of classes:          & \textbf{32} & \textbf{16} & \textbf{8} & \textbf{4-easy}     & \textbf{4hard-a} & \textbf{4hard-b} & \textbf{4hard-c} & \textbf{4hard-d} \\ \hline
Cross-Entropy Loss          & 58.1 (0.7)  & 68.8 (0.4)  & 78.0 (0.6) & \textbf{89.2 (0.3)} & 56.1 (0.5)       & 63.2 (0.9)       & 54.3 (1.0)       & 67.4 (0.6)       \\
Supervised Contrastive Loss & 58.6 (0.5)  & 67.9 (0.6)  & 77.0 (0.8) & 88.8 (0.5)          & 55.4 (0.5)       & 63.7 (1.1)       & 53.3 (0.8)       & 68.1 (0.7)       \\
Label-aware Contrastive Loss &
  \textbf{60.1 (0.2)} &
  \textbf{69.6 (0.5)} &
  \textbf{78.7 (0.4)} &
  88.8 (0.6) &
  \textbf{57.5 (0.7)} &
  \textbf{64.2 (0.7)} &
  \textbf{55.6 (0.6)} &
  \textbf{69.5 (0.5)}
\end{tabular}
}
\caption{Case study using class subsets of EmpatheticDialogues. For brevity, we only report accuracy scores. 
Column headers give the number of class labels in that comparison. 4-easy denotes a coarse-grained set of four emotions that are more easily distinguishable (on which we predicted that LCL would not add much), while the 4-hard sets denote fine-grained sets of four emotions that are semantically more similar. 
Results shown are averaged over 10 runs, with standard deviations in parentheses.}
\label{tab:fine-emotion}
\end{table*}

We designed LCL to increase inter-class contrast, and we see marked improvements for all the tasks studied except for the 2-class (SST-2) classification. We hypothesized that LCL should do better with an increasing number of classes, but unfortunately it is difficult to draw that inference from Tables \ref{tab:emo} and \ref{tab:sentiment-results} as each dataset only provides one datapoint about number of classes, and there are also differences across datasets which is difficult to control for. Thus, in this experiment, we used the dataset with the largest number of emotion classes, Empathetic Dialogues (with 32-classes), and subsampled some fraction of emotion classes from this dataset to create ``mini-datasets'' of differing number of emotion classes. This allows us to systematically vary the number of classes that our LCL-tuned model has to learn to classify, and examine the performance of the model. We predict that LCL will have a greater contribution to performance when (i) the number of classes is larger, and (ii) the classes are more confusable. 


The full dataset has 32-classes. We randomly sampled a partition of 16 emotions\footnote{
\{\emph{Afraid}, \emph{Angry}, \emph{Annoyed}, \emph{Anxious}, \emph{Confident}, \emph{Disappointed}, \emph{Disgusted}, \emph{Excited}, \emph{Grateful}, \emph{Hopeful}, \emph{Impressed}, \emph{Lonely}, \emph{Proud}, \emph{Sad}, \emph{Surprised}, \emph{Terrified}\}}, 
and 8 emotions\footnote{
\{\emph{Angry}, \emph{Afraid}, \emph{Ashamed}, \emph{Disgusted}, \emph{Guilty}, \emph{Proud}, \emph{Sad}, \emph{Surprised}\}}.
We also created several subsets of 4-emotions. We designed a ``4-easy'' with 4 widely separated emotion classes (4-easy: \{\emph{Angry}, \emph{Afraid}, \emph{Joyful}, \emph{Sad}\}) which are the same classes as EmoInt and comprise a subset of \citet{ekman1999basic}'s list of six ``basic" emotions. (We predicted that LCL would not perform too well on this easy subset). 

We adopted a data-driven approach to pick the ``hard" subsets by picking the most-confusable sets of 4 emotions. First, we trained a standard cross-entropy loss model (similar to our weighting network in LCL in Fig.\ref{fig:model}), to obtain the 32-by-32 confusion matrix, which gives us an estimate of how confusable each pair of classes is. We exhaustively enumerated all 35,960 (32-choose-4) 4-class combinations: For each combination we extracted the corresponding 4x4 sub-matrix of the 32-by-32 confusion matrix, and calculated the sum of the off-diagonal elements of the 4x4 sub-matrix. The highest confusable combination of emotions was (4-hard-a: \{\emph{Anxious}, \emph{Apprehensive}, \emph{Afraid}, \emph{Terrified}\}). After excluding these emotions, the next-most confusable combinations were (4-hard-b: \{\emph{Devastated}, \emph{Nostalgic}, \emph{Sad}, \emph{Sentimental}\}), (4-hard-c: \{\emph{Angry}, \emph{Ashamed}, \emph{Furious}, \emph{Guilty}\}), and (4-hard-d: \{\emph{Anticipating}, \emph{Excited}, \emph{Hopeful}, \emph{Guilty}\}). We predicted that for all of these ``hard" sets that contain confusable emotions, LCL should outperform the other methods.


The results from this case study are given in Table \ref{tab:fine-emotion}. For the 32, 16, and 8-class classification, as we predicted, we see a robust and consistent improvement of our proposed LCL over SCL and cross-entropy loss (16 classes: LCL$>$SCL, $t=6.28, p<.001$; LCL$>$CEL, $t=3.82, p=.001$; 8 classes: LCL$>$SCL, $t=6.27, p<.001$; LCL$>$CEL, $t=3.16, p=.007$). For the easy 4-class classification where the classes are conceptually ``far apart'', and hence, contrastive learning should not add much, we see that all three methods perform identically well ($p>.15$). But when we consider the more difficult 4-class classifications where the classes are much more conceptually similar, then LCL outperforms the other two methods by a statistically-significant margin (all $p$'s $<.05$ except for LCL and SCL in 4-hard-b because of the high SD's in that comparison). Thus, our results provide evidence that LCL is an effective fine-tuning strategy, especially when there are a large number of highly-similar classes.

\subsection{Quantifying model confidence}

Finally, we wanted to try to quantify the intuition that LCL helps to reduce the confusion among confusable classes. Beyond looking at the top-1 accuracy, we turned to the distribution of prediction scores among the different emotion classes. If LCL helps the model to better differentiate emotion classes, then we should also see this in the distribution of prediction scores for the different classes. For example, consider an example where \emph{devastated} is the model's predicted label, and \emph{sad} is a closely confusable class; if LCL helps to sharpen the model's ability to differentiate closely confusable classes, then the model's prediction score for \emph{devastated} should also be much higher than that for \emph{sad}. In general, we predict that LCL would result in more ``peaky'' distributions.

\begin{figure}[htb]
\centering
\includegraphics[width=\columnwidth]{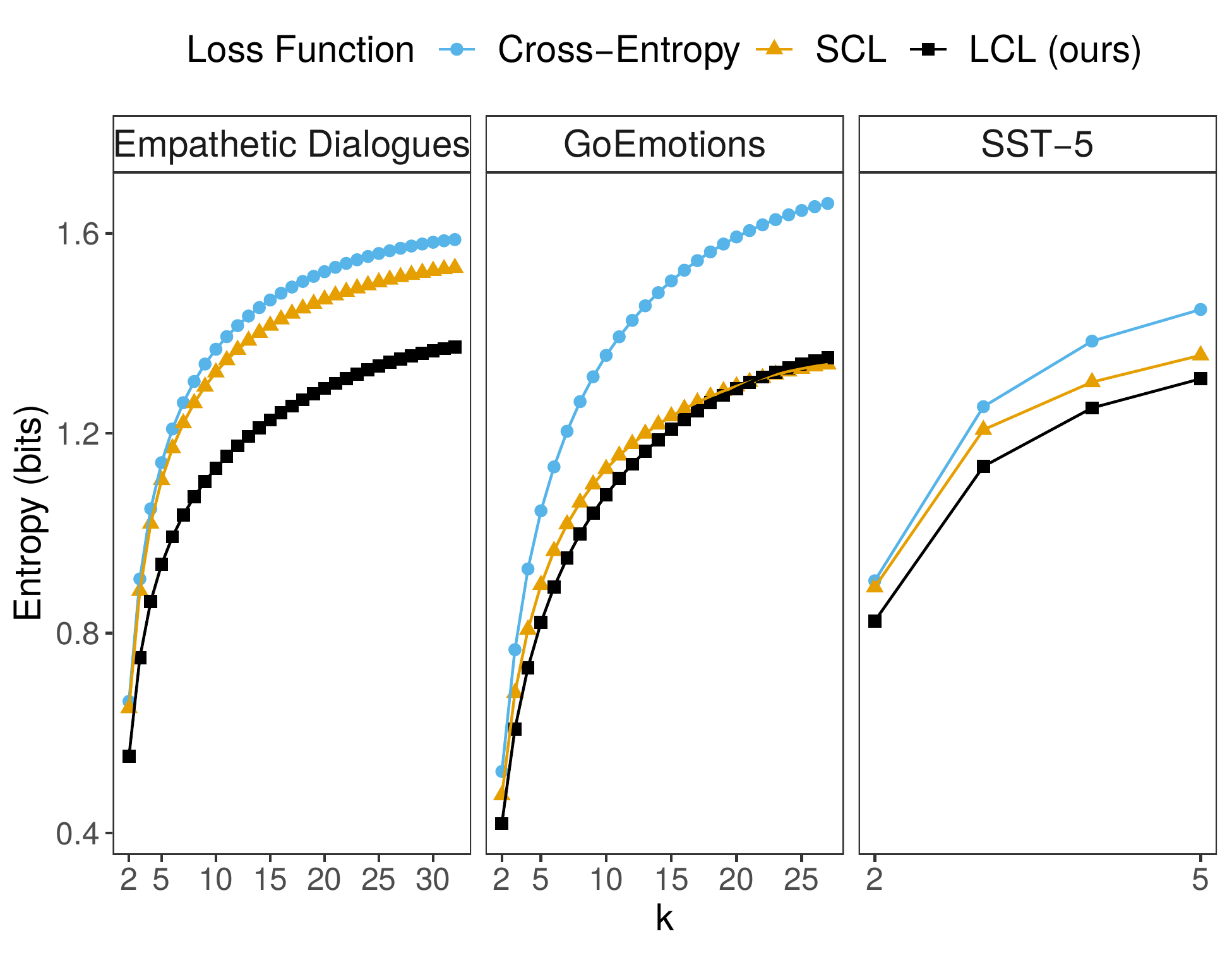}
\caption{Averaged entropy of the prediction score distributions, for the top-$k$ choices. Here, decreasing entropy carries the intuition that the distribution is more ``peaky'', such that the model is less confused by close alternatives.}
\label{fig:entropy}
\end{figure}

We propose to use information-theoretic entropy to quantify this. We predict that LCL would result in prediction score distributions with lower entropy, which correponds to more ``peaky'' distributions. For a data point $x_i$, let us denote the prediction score as $S \in \mathbb{R}^{C}$, where $C$ is the total number of class labels. We then take the top-$k$ prediction scores $S_k$ as the sub-vector of $S$ with the $k$-largest values (i.e., for $k=2$, $S_k$ would consist of the two largest values in $S$). We normalize $S_k$ to sum to 1, and then calculate the entropy:
\begin{equation}
 \text{Entropy}_{k} = - \sum_{k} s_{k} \cdot \log_{2}(s_{k})
\end{equation}

In Figure \ref{fig:entropy}, we present the averaged entropy of our model's prediction scores, plotted against $k$ for the fine-grained emotion classification (Empathetic Dialogues and GoEmotion) and fine-grained sentiment analysis (SST-5). For Empathetic Dialogues, we see that LCL produces distributions with far lower entropies, compared to cross-entropy and SCL, and this is true as we look across the top-$k$ classes. For GoEmotions, we see a slightly different pattern, where both SCL and LCL produce markedly less-entropic distributions compared to the vanilla cross-entropy loss, but there was not much difference between SCL and our LCL. Finally, for SST-5, which was the most fine-grained sentiment analysis task we looked at, we start to see the same pattern that LCL produces the lowest entropy distributions, but this inference is limited by the small domain of $k$.

This post-hoc analysis suggests that LCL helps the model to learn prediction distributions that are more confident. Note that this analysis looks at the confidence of the model's choice compared to the space of possible choices, and is independent of whether or not the predictions are correct (i.e., an inaccurate but confident model will also produce peaky, lower-entropic distributions), and so this result complements the other evaluation metrics used (accuracy and F1-scores).

\section{Conclusion}

In this paper we introduced a Label-aware Contrastive Loss that weights (negative) classes based on how closely confusable they are with the target class. Fine-tuning with LCL showed increased classification performance, especially in situations with (i) larger number of classes, and (ii) more confusable classes. LCL also seems to encourage the model to be more confident in its decisions.

We view our approach as just one way to instantiate the general idea of adaptively weighting different classes, and future work could explore other methods such as incorporating external knowledge about the class labels, or incorporating different distance metrics between different classes. We feel that this class of approaches are promising, as they exemplify the idea that not all negative classes are or should be treated equally.


\section*{Acknowledgements}

This research is supported by the National Research Foundation, Singapore under its AI Singapore Program (AISG Award No: AISG2-RP-2020-016).

\bibliography{anthology,custom}
\bibliographystyle{acl_natbib}
\clearpage
\appendix

\section{Appendix}
\label{sec:appendix}

\subsection{Evaluation metrics}
We use top-1 accuracy and weighted macro F1-score. Weighted F1-score takes care of the imbalance in the label distribution and the equation for weighted macro F1-score is given by,
\begin{equation}
    \text{weighted F1} = 2 \sum_{c} \frac{n_{c}}{N} \frac{precision_{c} \times recall_{c}}{precision_{c} + recall_{c}}
\end{equation}
where, $n_{c}$ is number of samples in class $c$ and $N$ is the total number of samples.
\section{Experiment settings}

For fine-tuning pre-trained models using Label-aware Contrastive Loss (LCL), we use Adam optimiser with $\beta_{1}$ set to 0.9, $\beta_{2}$ set to 0.999 and $\epsilon$ set to 1e-06 with weight decay set to 1e-02. We used manual search for hyper-parameter search and the best model was chosen based on the best top-1 accuracy yielded in the validation data. Learning rate was chosen from set \{1e-05, 2e-05, 3e-05\}, loss scaling factor $\alpha$ was chosen from $\{0.1,0.2, \cdots ,0.5\}$ and temperature parameter $\tau$ was chosen from the set $\{0.1,0.3,0.5\}$. The best parameter setting of LCL are as follows, for EmpatheticDialogues, EmoInt, SST-5, SST-2, GoEmotions learning rate was found to be 2e-05 and for ISEAR it was found to be 3e-05. The $\alpha$ setting was found to be 0.5 for EmpatheticDialogues, EmoInt, SST-5, SST-2, ISEAR and 0.1 for GoEmotions. For all datasets except SST-5 the temperature parameter was found to be 0.3 and for SST-5 it was found to be 0.1. Batch size was set to 10 for all the datasets, as we have one augmented sample for every input sample the effective batch-size becomes 20.

For EmoInt the tweet data was cleaned using by removing non-ascii  characters,  letter  repetitions  and extra white-spaces. In addition, all the user-mentions and links were replaced to unique identifiers. We ran all our experiments using machine equipped with a NVIDIA Tesla T4 GPU.

\subsection{Average runtime and parameters}
During training time, the number of parameters trainable parameters is the combined number of parameters of the primary encoder and the weighting network, in our case we use the base of ELECTRA for both which has ~ 110M parameters. The average run-time of the model for one epoch was found to be 2.9 min for EmoInt , 5.2 min for ISEAR, 19.8 min for GoEmotions, 19.7 min for EmpatheticDialogues, 6.1 min for SST-2 and 8.2 min for SST-5. 
\section{Validation performance}
The corresponding validation performance for the reported test results are provided for emotion classification task in Table \ref{tab:val_emo} and sentiment analysis task in Table \ref{tab:val_sentiment}.
\begin{table}[h]
\resizebox{0.5\textwidth}{!}{%
\begin{tabular}{lcl}
\hline
& SST-2 & \multicolumn{1}{c}{SST-5} \\
& Acc / \% & \multicolumn{1}{c}{Acc / \%} \\ \hline
Cross-Entropy & 94.2 (0.4)          & 53.3 (0.7)                   \\
SCL \cite{gunel2021supervised}       & 94.4 (0.1)          & 54.5 (1.2)                   \\
LCL           & \textbf{94.8 (0.2)} & \textbf{55.4 (0.8)}   \\ \hline 
\end{tabular}%
}
\caption{Summary of validation results for sentiment analysis task. The results shown are averaged over 5 runs and the standard deviation is provided in the brackets.}
\label{tab:val_sentiment}
\end{table}

\begin{table*}[]
\resizebox{\textwidth}{!}{%
\begin{tabular}{lccll|llll}
\hline
Dataset : &
  \multicolumn{2}{c}{Empathetic Dialogues} &
  \multicolumn{2}{c|}{GoEmotions} &
  \multicolumn{2}{c}{ISEAR} &
  \multicolumn{2}{c}{EmoInt} \\
Number of classes : & \multicolumn{2}{c}{32}  & \multicolumn{2}{c|}{27} & \multicolumn{2}{c}{7}   & \multicolumn{2}{c}{4}   \\
 &
  Acc / \% &
  F1 &
  \multicolumn{1}{c}{Acc / \%} &
  \multicolumn{1}{c|}{F1} &
  \multicolumn{1}{c}{Acc / \%} &
  \multicolumn{1}{c}{F1} &
  \multicolumn{1}{c}{Acc / \%} &
  \multicolumn{1}{c}{F1} \\ \hline
Cross-Entropy       & 59.0 (0.2) & 58.1 (0.4) & 66.2 (0.2) & 65.3 (0.3) & 71.7 (0.4) & 71.7 (0.4) & 87.1 (1.0) & 87.1 (1.0) \\
SCL  \cite{gunel2021supervised}                & 58.9 (0.7) & 57.8 (0.8) & 64.9 (0.3) & 63.7 (0.3) & 72.2 (0.7) & 72.2 (0.7) & 87.9 (0.5) & 87.9 (0.5) \\
LCL &
  \textbf{60.3 (0.4)} &
  \textbf{59.7 (0.4)} &
  \textbf{66.0 (0.2)} &
  \textbf{65.3 (0.2)} &
  \textbf{72.6 (0.2)} &
  \textbf{72.6 (0.2)} &
  \textbf{88.8 (0.8)} &
  \textbf{88.9 (0.8)} \\ \hline
\end{tabular}%
}
\caption{Summary of validation results for emotion classification task. The results shown are averaged over 5 runs and the standard deviation is provided in the brackets.}
\label{tab:val_emo}
\end{table*}
\end{document}